\documentclass{article}
\usepackage{iclr2027_conference,times}
\usepackage[T1]{fontenc}
\usepackage[utf8]{inputenc}
\usepackage{amsmath,amssymb,mathtools}
\usepackage{graphicx,booktabs,array,tabularx,multirow}
\usepackage{algorithm,algpseudocode}
\usepackage{microtype,placeins}
\usepackage{url}
\usepackage[hidelinks]{hyperref}
\usepackage{makecell}
\usepackage{capt-of}

\newcommand{\X}[1]{\ifmmode\mathbf{X}\else\textbf{X}\fi}

\newcolumntype{Y}{>{\raggedright\arraybackslash}X}
\newcommand{\smallhead}[1]{\noindent\textbf{#1}}

\title{DORA: Dynamic Online Reinforcement Agent for Token Pruning in Vision Transformers}

\author{%
Kaixuan He$^{1,2}$ \quad
Song Chen$^{1}$ \quad
Yi Kang$^{1}$\thanks{Corresponding author.}
\\[2mm]
$^{1}$University of Science and Technology of China, Hefei, China
\\
$^{2}$Institute of Artificial Intelligence, Hefei Comprehensive National Science Center, Hefei, China
\\[1mm]
\texttt{hekaix@mail.ustc.edu.cn}
\quad
\texttt{\{songch,ykang\}@ustc.edu.cn}
}

\iclrfinalcopy

\begin{document}

\maketitle

\lhead{Preprint}

\begin{abstract}
Vision Transformers (ViTs) incur quadratic self-attention cost in the
number of tokens. Most token-reduction methods adapt token identities
within a prescribed layer-wise compression schedule, or search a static
mask offline, and thus limit online adaptation of when and how much to prune.
We propose \textbf{DORA} (\textbf{D}ynamic
\textbf{O}nline \textbf{R}einforcement \textbf{A}gent), which learns an
input-adaptive pruning policy itself for frozen ViTs. At each eligible
block, a hierarchical actor decides whether to prune, how many tokens
to remove, and which tokens to remove from each image's evolving
representation. Because early deletions change the states observed by
later decisions, DORA formulates pruning as a finite-horizon Markov
decision process. Complete-prefix shadow evaluations convert
final-prediction fidelity into localized per-step credit, while
closed-loop accuracy feedback adjusts the fidelity penalty toward a
shared accuracy-drop target.
A privileged critic and all shadow computations are training-only.
Deployment retains the frozen backbone and a lightweight actor that
applies hard deletion and packed variable-length FlashAttention,
converting token reduction into measured speedups. On ImageNet-1K with 
DeiT-Base, DORA reduces FLOPs by 38.4\% relative to the uncompressed 
backbone within one percentage point of accuracy loss. 
Averaged across four ViT-type backbones at matched accuracy,
DORA uses 13.2\% fewer FLOPs and achieves 32.4\% higher throughput
than the corresponding per-backbone baseline means. Under zero-shot
transfer to ImageNet-A, these gains widen to 20.3\% and 45.6\%,
respectively.
\end{abstract}

\section{Introduction}

Vision Transformers (ViTs)~\citep{dosovitskiy2021vit} incur quadratic
self-attention cost in the number of tokens, making token pruning or
merging a direct route to lower inference
cost~\citep{bolya2023tome,wang2024zerotprune}. Most existing methods
improve how tokens are scored, matched, or aggregated within a
prescribed layer-wise compression schedule: token identities may depend
on the image, but the per-layer budget is shared across
inputs~\citep{bolya2023tome,wang2024zerotprune,lee2024mctf,tran2024pitome,choi2025representationshift}.
This restriction is mismatched to input-dependent redundancy: a common
schedule can retain unnecessary tokens for one image and remove useful
detail from another. We argue that the compression budget itself should
be input-adaptive, decided online from each image's evolving
representations.

Token pruning is also not a sequence of independent local ranking
problems. An early deletion saves more downstream computation, but it
changes the representations on which later decisions are made; tokens
that are individually expendable may be unsafe to remove together. We
therefore treat pruning as online sequential control, formulated as a
finite-horizon Markov decision process (MDP). This formulation lets
the objective account for the long-term effect of discrete pruning
actions rather than relying only on local attention or similarity
proxies. V-Pruner also uses PPO, but searches for an offline mask shared
across inputs~\citep{yao2026vpruner}; our learned policy remains active
at inference.

\begin{figure}[t]
    \centering
    \includegraphics[width=\linewidth]{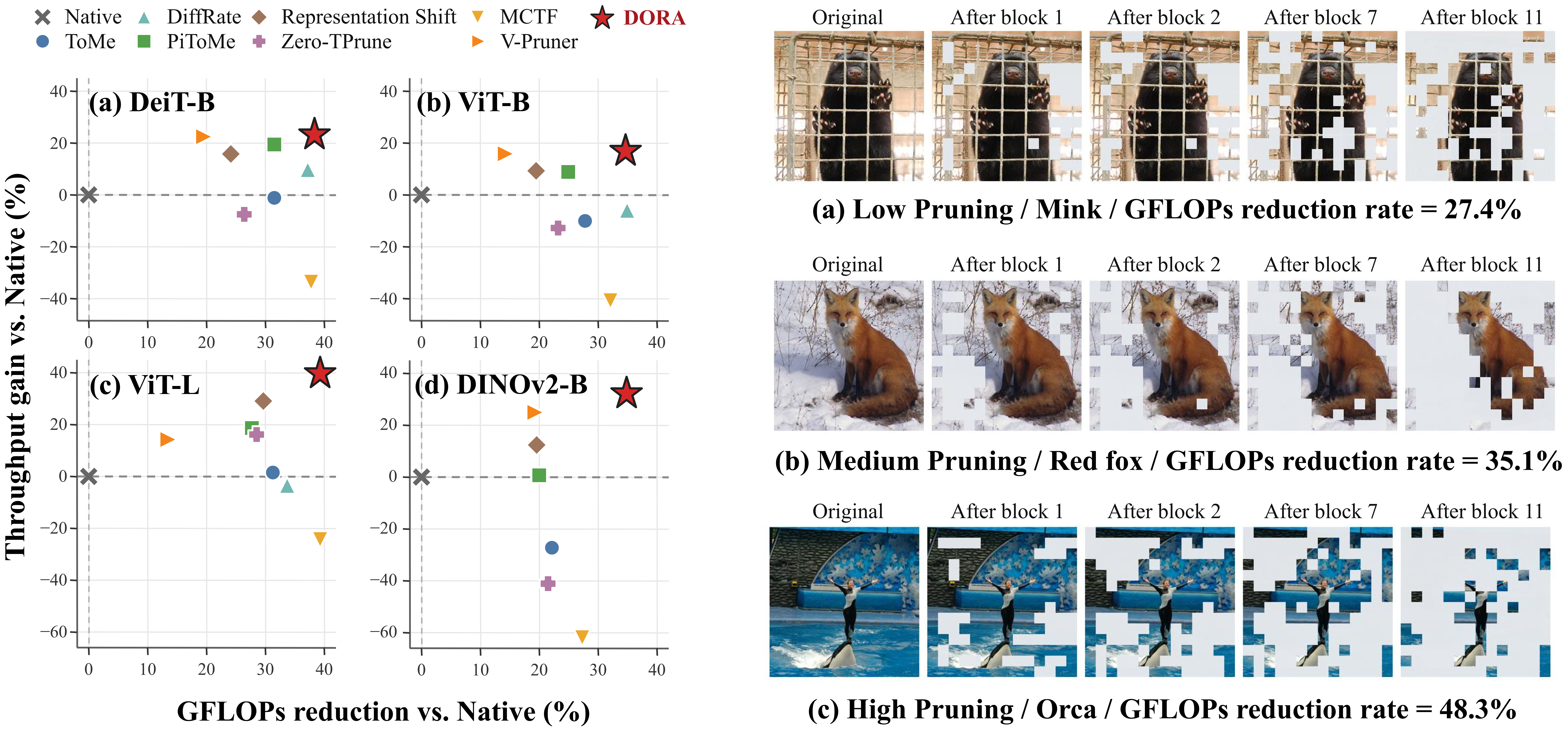}
    \caption{\textbf{Joint efficiency and input-dependent behavior.}
    Left: GFLOPs reduction and measured throughput gain at
    accuracy-matched operating points across four ViT-type backbones;
    DORA (star) achieves the strongest joint efficiency at comparable
    accuracy. Right: representative DeiT-B trajectories showing that the
    learned policy applies different pruning intensities and retains
    different spatial evidence across images.}
    \label{fig:main_dora}
\end{figure}

We introduce \textbf{DORA} (\textbf{D}ynamic \textbf{O}nline
\textbf{R}einforcement \textbf{A}gent), an RL framework that learns the
pruning policy itself for frozen ViTs. A hierarchical actor decides
\emph{where} to prune, \emph{how many} tokens to remove, and \emph{which}
tokens to remove (Figure~\ref{fig:main_dora}, right), with rollouts that physically delete tokens so that
later decisions observe the state created by earlier actions. Two
training-only mechanisms make this policy learnable without
backbone-specific manual tuning: complete-prefix shadow evaluations
convert final-prediction fidelity into localized credit for individual
pruning actions, and closed-loop accuracy feedback adjusts the fidelity
penalty toward a single accuracy-drop target shared across backbones.
Deployment retains only the frozen backbone and a lightweight
deterministic actor, whose hard deletion and packed variable-length
attention turn token reduction into measured wall-clock speedups. At
matched ImageNet-1K accuracy, DORA averages 13.2\% fewer FLOPs and
32.4\% higher throughput than the baseline means across four ViT-type
backbones (Figure~\ref{fig:main_dora}, left); under zero-shot transfer to ImageNet-A, these gains widen to
20.3\% and 45.6\% (Table~\ref{tab:main_efficiency_joint}).

Our contributions are:
\begin{itemize}
    \item We formulate dynamic online token pruning as a finite-horizon MDP and propose a hierarchical policy consisting of a gate, a budget head, and a selector, which jointly decide where to prune, how much to prune, and which tokens to remove for each input.
    
    \item We introduce localized credit assignment based on complete-prefix shadow evaluations, together with closed-loop accuracy feedback for regulating the compression--fidelity trade-off, eliminating backbone-specific manual tuning.

    \item We couple adaptive hard pruning with packed variable-length FlashAttention and avoid attention-map materialization, token-size corrections, and batch-wide padding, all contributing to higher end-to-end throughput.

\end{itemize}
\section{Related Work}

\smallhead{Token pruning and merging.}
Token reduction shortens the sequence processed by a ViT through
either pruning or merging. Pruning methods score and discard tokens
using attention- or graph-based importance: ATS samples tokens from
classification-token attention scores without additional
parameters, keeping a per-image varying number of tokens up to a preset
maximum~\citep{fayyaz2022ats}, while Zero-TPrune builds an
attention graph and ranks tokens with Weighted
PageRank~\citep{wang2024zerotprune}. Merging methods instead
aggregate tokens: ToMe matches keys via bipartite soft
matching~\citep{bolya2023tome}, ATC clusters tokens
bottom-up~\citep{haurum2024atc}, MCTF combines similarity,
informativeness, and token size with one-step-ahead
attention~\citep{lee2024mctf}, and PiToMe protects informative
tokens with a graph-energy criterion before
matching~\citep{tran2024pitome}. These methods improve token
scoring, matching, or aggregation; yet while the compressed tokens
may be chosen per image, the per-layer budget is prescribed and
shared across inputs.

\smallhead{Adaptive token reduction.}
DynamicViT predicts image-conditioned token masks but keeps a
predefined token count at each sparsification stage and fine-tunes
the backbone together with the prediction
modules~\citep{rao2021dynamicvit}. DiffRate learns layer-wise
compression rates under a computation constraint, but the learned
rates are shared across images~\citep{chen2023diffrate}. RL4EViT
formulates token pruning as a Markov game and uses MAPPO to learn
input-dependent binary keep-or-prune decisions for individual
tokens, but the pruning stages remain manually predefined and
pruned tokens are suppressed through attention masking rather than
physically removed from the sequence~\citep{lu2025rl4evit}.
V-Pruner applies PPO with a Fisher-information prior, but as an
offline mask search whose pattern is shared across
inputs~\citep{yao2026vpruner}. Across these methods, some element of the schedule---the stages, the token counts, or the mask pattern---remains fixed and shared across
inputs.

\smallhead{Execution with fused attention.}
FlashAttention computes attention without materializing the full
probability matrix~\citep{dao2022flashattention}, so methods that
rank tokens by intermediate attention probabilities---ATS,
Zero-TPrune, and MCTF among them---cannot use the standard fused
operator
directly~\citep{fayyaz2022ats,wang2024zerotprune,lee2024mctf}.
Representation Shift restores compatibility with a training-free
importance score derived from representation
changes~\citep{choi2025representationshift}. Merging methods that
add token-size terms to the attention logits, such as proportional
attention, likewise require modifications to the
operator~\citep{bolya2023tome,haurum2024atc}.
\section{Methodology}
\label{sec:method}

\begin{figure}[t]
    \centering
    \includegraphics[width=\linewidth]{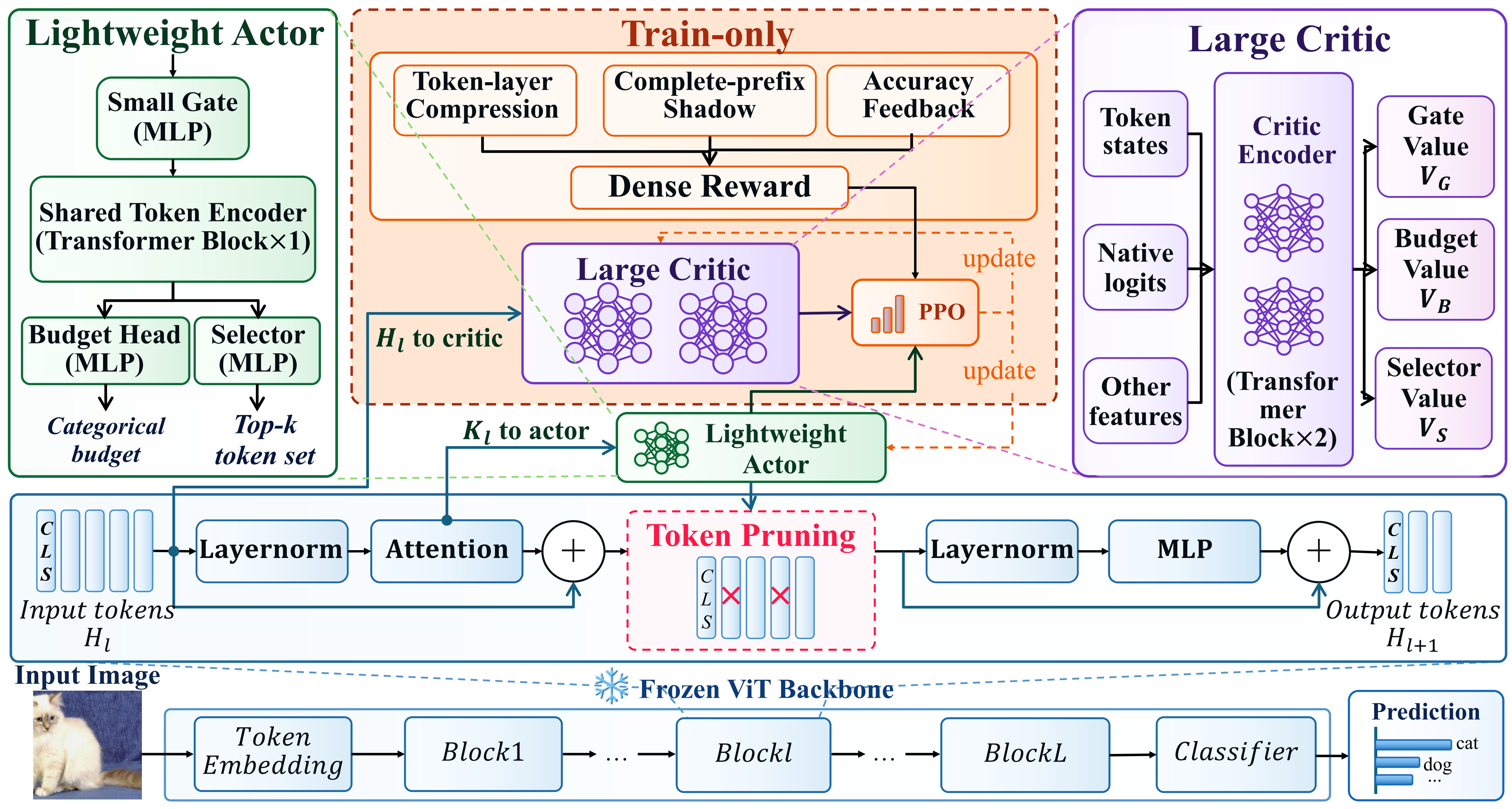}
    \caption{\textbf{Training--deployment separation in DORA.}
    Training uses a hierarchical actor with complete-prefix rewards,
    accuracy feedback, and a privileged critic; deployment retains only
    the frozen ViT and deterministic actor with hard pruning and
    variable-length FlashAttention.}
    \label{fig:dora_architecture}
\end{figure}

\subsection{Overview and formulation}
\label{sec:method-overview}

DORA addresses the limitations above with a single learned policy:
the compression budget is decided per image rather than prescribed,
each decision accounts for its downstream effects through sequential
rollouts, and hard deletion keeps execution compatible with the
unmodified fused attention operator. Figure~\ref{fig:dora_architecture}
gives an overview. A lightweight hierarchical actor is attached to a
frozen ViT: its gate decides whether to prune, its budget head
determines how many tokens to remove, and its selector chooses which
tokens to remove. 

Consider a pretrained ViT with $L$ Transformer blocks. At block $l$,
let
$H_l=[x_{l,\mathrm{CLS}};x_{l,1},\ldots,x_{l,N_l}]
\in\mathbb{R}^{(N_l+1)\times d}$,
where $x_{l,\mathrm{CLS}}$ is the CLS classification-token
representation, $N_l$ is the number of surviving visual tokens, and
$d$ is the token dimension. $N_0$ denotes the initial visual-token
count. DORA may prune at any $l\in\{0,\ldots,L-1\}$, while CLS is
always retained.

\smallhead{Pruning operator.}
The actor makes its decision from the current attention keys, while
deletion takes effect after the attention residual:
\begin{equation}
\begin{aligned}
H_l^{\mathrm{attn}}
&=H_l+\operatorname{Attention}_l(
\operatorname{LayerNorm}_l(H_l)),\\
H_{l+1}
&=H_l^{\mathrm{keep}}+
\operatorname{MLP}_l(
\operatorname{LayerNorm}'_l(H_l^{\mathrm{keep}})).
\end{aligned}
\label{eq:pruning-operator}
\end{equation}
Here $H_l^{\mathrm{keep}}$ removes the selected visual-token rows
while preserving CLS and survivor order. If $k_l$ tokens are removed,
$N_{l+1}=N_l-k_l$, with $k_l=0$ when the gate is off. Removed tokens
therefore participate in the current attention but not in the current
MLP or subsequent blocks; the sequence is physically shortened
without merging or zero masking.

\smallhead{MDP formulation.}
We formulate dynamic online token pruning as a finite-horizon MDP
$\mathcal{M}=(\mathcal{S},\mathcal{A},\mathcal{P},\mathcal{R},\gamma)$
with one decision step per block and $\gamma=1$. At block $l$, the
policy observes $o_l=(K_l,h_l,l)$, where $K_l$ contains the surviving
attention keys and $h_l$ summarizes the pruning history. Its
hierarchical action $a_l=(g_l,k_l,S_l)$ specifies whether to prune,
how many tokens to remove, and their indices. The transition is
deterministic under the frozen backbone, while the reward balances
token-layer compression and incremental prediction-fidelity loss
(Section~\ref{sec:method-reward}).

DORA learns a policy $\pi$ that maximizes the undiscounted return
$J(\pi)=\mathbb{E}_{\pi}\!\left[\sum_{l=0}^{L-1} r_l\right]$, so that
intervention locations, pruning budgets, and token identities adapt to
each image's evolving representation.

\subsection{Hierarchical policy}
\label{sec:method-policy}

The actor consists of a gate followed by a budget--selector
controller that shares a token encoder. All layer widths and
embedding sizes are listed in Appendix~\ref{app:actor-architecture}.

\smallhead{Gate.}
The gate reads only the CLS key of $K_l$, concatenated with $h_l$
and a learned block embedding, and predicts an activation
probability $p_l^G$ through a small MLP. When $g_l=0$, no tokens are
removed and the controller is skipped entirely; the gate is forced
off at blocks where no positive removal count is legal. Reading only
the CLS key keeps the gate's cost negligible relative to the block
it guards.

\smallhead{Shared encoder.}
When the gate is active, the controller projects all surviving keys,
adds a learned block embedding and the pruning history $h_l$, and
processes the sequence with a single lightweight Transformer block,
producing $Z_l\in\mathbb{R}^{(N_l+1)\times128}$. This encoder models relations among surviving tokens, so scores can depend on the context shaped by preceding pruning decisions.

\smallhead{Budget head and selector.}
A budget head maps the encoded CLS token to a categorical
distribution $\pi_B(\cdot\mid o_l)$ over a backbone-specific grid of
candidate removal counts (Appendix~\ref{app:actor-budgets}), masking
counts that are invalid at the current block before the softmax.

Given the chosen budget $k_l$, the selector scores each visual token
for removal:
\begin{equation}
    s_{l,j}=\operatorname{MLP}_{S}\!\left(
        Z_{l,j}+W_k\frac{k_l}{N_l}
    \right),\qquad j=1,\ldots,N_l.
    \label{eq:selector}
\end{equation}
Where $Z_{l,j}$ is the encoded representation of token $j$ and $W_k$
is a learned projection of the removal fraction. A higher score
indicates a stronger preference for removal; the CLS token is never
selected. Conditioning on $k_l/N_l$ allows token rankings to change
with the requested pruning intensity.

\smallhead{Sampling and deployment.}
During training, the gate and budget are sampled from Bernoulli and
categorical distributions, and exactly $k_l$ distinct tokens are
sampled without replacement using an ordered Plackett--Luce
distribution over the scores~\citep{plackett1975permutations}.
Deployment uses a gate threshold of $0.5$, the largest budget logit,
and the $k_l$ highest selector scores. 

\subsection{Localized rewards from complete-prefix evaluation}
\label{sec:method-reward}

To localize prediction degradation, we ask, for every block $l$ in a
sampled rollout: what would the backbone predict if pruning stopped
right after block $l$? We answer by replaying the recorded actions at
blocks $0$ through $l$ and running all remaining blocks with no further
pruning, producing logits $z^{(l)}$ from the original classifier. We
call each such replay a \emph{complete-prefix shadow evaluation}, as it
branches off the realized trajectory without affecting it. Its fidelity
cost relative to the unpruned logits $z^{\mathrm{N}}$ uses a
temperature-scaled soft-target KL divergence with the conventional
$\tau^2$ scaling from knowledge distillation
\citep{hinton2015distilling}:
\begin{equation}
    D_l=\tau^2\operatorname{KL}\!\left(
    \operatorname{softmax}(z^{\mathrm N}/\tau)
    \,\|\,
    \operatorname{softmax}(z^{(l)}/\tau)\right),\qquad \tau=3.
    \label{eq:kd}
\end{equation}
Because consecutive prefixes share all earlier actions and contain no
later ones, $D_l-D_{l-1}$ attributes the change in final-prediction
fidelity to the action at block $l$ alone. The difference is signed: positive when the action moves the prediction away from the reference, negative when it moves closer.

Rewarding fidelity alone would discourage pruning altogether, so we
pair it with a compression term measured across the whole backbone,
since removing a token earlier saves computation in more downstream
blocks. Let $N_0$ denote the initial number of visual
tokens and $N_l^{\mathrm{post}}$ the number remaining after pruning at
block $l$. The token-layer compression, and the increment we assign to block $l$,
are
\begin{equation}
    C=\frac{1}{LN_0}\sum_{l=0}^{L-1}\left(N_0-N_l^{\mathrm{post}}\right),
    \qquad
    \Delta C_l=\frac{k_l(L-l)}{LN_0},
    \label{eq:compression}
\end{equation}
so that $C=\sum_{l=0}^{L-1}\Delta C_l$: each removed token contributes
in proportion to the number of blocks it skips.

Combining the two terms, the reward at training update $u$ is
\begin{equation}
    r_l=25\,\Delta C_l-a_u\left(D_l-D_{l-1}\right),
    \qquad D_{-1}=0,
    \label{eq:reward}
\end{equation}
where the shared factor $25$ sets the scale of the compression term relative to
the fidelity cost, and $a_u$ is the fidelity coefficient adjusted by accuracy feedback (Section~\ref{sec:method-feedback}). The incremental fidelity terms telescope to the final trajectory-level divergence, distributing the end-to-end cost across decisions; we add no separate terminal or classification-event reward. The derivation and shadow-evaluation procedure are provided in Appendix~\ref{app:reward}.

\subsection{Closed-loop accuracy feedback}
\label{sec:method-feedback}

A fixed fidelity coefficient does not produce the same quality regime
across backbones. DORA therefore evaluates the deterministic deployment
policy periodically on a held-out shard of the training set. Let $d_u$
be its paired Top-1 drop relative to the Native model on that shard,
i.e., degraded minus improved prediction fractions. The coefficient is updated toward a one-percentage-point target:
\begin{equation}
    a_{u+1}=\operatorname{clip}\!\left(
        a_u+5\,\operatorname{clip}\!\left(
        \frac{d_u-0.01}{0.01},-2,2\right),0,300\right).
    \label{eq:feedback}
\end{equation}
The penalty increases when the measured drop exceeds the target and is
relaxed otherwise; labels are used only to compute feedback accuracy,
and the target regulates training without being a hard constraint on
unseen validation data. Cadence, shard construction, and
initialization are given in Appendix~\ref{app:feedback}.

\subsection{Optimization and inference}
\label{sec:method-inference}

\smallhead{Training-only critic.}
A separate, larger critic is available only during training. Beyond the
current token state, it reads privileged information unavailable to the
actor: the Native logits of the same image and scalar features
describing the block position, token count, pruning history, reference
confidence, and the current fidelity coefficient
(Appendix~\ref{app:critic-architecture}). It predicts value baselines for the gate,
budget, and selector decisions; the gate and budget values are computed
without the current removal count, while the selector value
additionally receives the chosen budget and is evaluated before token
selection, so no baseline observes which tokens will be removed or the
resulting prediction.

\smallhead{Policy optimization.}
We use GAE~\citep{schulman2015gae} and separate clipped PPO
objectives~\citep{schulman2017ppo}, one per decision type. A common
$\lambda$-return is shared by all three decision types, each
subtracting its own value baseline, and the advantages are normalized
separately over their eligible transitions. The critic
architecture, value targets, PCGrad handling~\citep{yu2020pcgrad}, and
hyperparameters are specified in Appendix~\ref{app:critic}.

\smallhead{Inference.}
At inference, only the frozen backbone and deterministic actor are
retained. Surviving sequences are compacted into a packed buffer with
explicit sequence boundaries that prevent attention between different
images, and processed with variable-length
FlashAttention~\citep{dao2022flashattention}. Compaction preserves per-image token order, and only gate-active examples enter the controller. The actor requests neither attention probabilities nor token-size corrections, so the backbone uses the
unmodified fused attention operator; reported throughput includes the
actor, token selection, packing, and compaction.
\section{Experiments}

\subsection{Experimental setup}
\label{sec:exp-setup}

\smallhead{Backbones.}
We evaluate four frozen pretrained backbones: DeiT-B/16, ViT-B/16 and
ViT-L/16 with AugReg checkpoints, and DINOv2-B/14 with its official
linear readout~\citep{touvron2021deit,steiner2022augreg,oquab2023dinov2}.
The patch-16 backbones process 196 visual tokens at
$224{\times}224$ resolution, while DINOv2-B/14 processes 256. All
methods under a backbone use the same weights and the checkpoint's
official evaluation preprocessing.

\smallhead{Data and checkpoint selection.}
DORA is trained only on mutually disjoint subsets of the ImageNet-1K
training set~\citep{russakovsky2014imagenet}; checkpoints are selected on a held-out development split
by minimizing GFLOPs subject to a one-percentage-point Native accuracy
drop. The DORA checkpoint is selected on the development split and frozen
before evaluation on ImageNet-1K validation and ImageNet-A.
Baseline operating points are matched to DORA's realized accuracy
under the protocol in Appendix~\ref{app:experiments-selection}.

\smallhead{Baselines.}
We compare with seven frozen-backbone token-reduction methods: ToMe,
PiToMe, MCTF, DiffRate, Representation Shift, Zero-TPrune, and
V-Pruner. For Zero-TPrune, we use a paper-based reimplementation.
For V-Pruner, the per-budget offline mask
search is repeated independently on each backbone; this search is an
offline cost and is excluded from deployment measurements.

\smallhead{Metrics and runtime measurement.}
All final numbers are produced under a single BF16 inference contract,
so each backbone has exactly one Native reference. We report full-set
Top-1 accuracy, complete-forward GFLOPs
($1\,\mathrm{MAC}=2\,\mathrm{FLOPs}$, including patch embedding and the
classifier), and GPU-resident end-to-end throughput on one RTX~4090 at
batch size~320. Timing includes each method's online selection, merging
or pruning, packing, and compaction overhead, and excludes data loading
and host-to-device transfer. DORA uses packed variable-length
FlashAttention; other methods use the fastest attention path compatible
with their semantics. Exact data splits, checkpoint selection,
implementations, and timing protocol are given in
Appendix~\ref{app:experiments}.

\subsection{Accuracy--compute--throughput trade-offs}
\label{sec:accuracy_compute_throughput}
\begin{table}[t]
\centering
\caption{Accuracy-matched comparison with seven token-reduction
baselines on ImageNet-1K and ImageNet-A. Baseline operating points
match DORA within 0.1 pp Top-1. \emph{Avg. baselines} is the unweighted
mean over available baselines; ``--'' denotes no matching point. Bold
marks the best efficiency value in each backbone block and the DORA row.}
\label{tab:main_efficiency_joint}
\normalsize
\setlength{\tabcolsep}{1.10pt}
\renewcommand{\arraystretch}{0.85}
\begin{tabular*}{\linewidth}{@{\extracolsep{\fill}}cccccccc@{}}
\toprule
\multirow{4}{*}{Backbone} &
\multirow{4}{*}{Method} &
\multicolumn{3}{c}{ImageNet-1K} &
\multicolumn{3}{c}{ImageNet-A} \\
\cmidrule(lr){3-5}\cmidrule(lr){6-8}
& &
\makecell{Top-1 Acc\\(\%)} &
\makecell{GFLOPs} &
\makecell{Throughput\\(images/s)} &
\makecell{Top-1 Acc\\(\%)} &
\makecell{GFLOPs} &
\makecell{Throughput\\(images/s)} \\
\midrule

\multirow{10}{*}{DeiT-B}
& Native        & 81.8 & 35.2 & 3230
                & 27.9 & 35.2 & 3230 \\
& ToMe          & 80.9 & 24.1 & 3163
                & 26.6 & 29.6 & 2433 \\
& DiffRate      & 80.8 & 22.1 & 3512
                & --   & --   & -- \\
& PiToMe        & 80.8 & 24.1 & 3821
                & 26.7 & 23.6 & 3844 \\
& MCTF          & 80.8 & 21.9 & 2122
                & 26.7 & 26.6 & 1647 \\
& Zero-TPrune   & 80.8 & 25.9 & 2960
                & 26.7 & 23.2 & 3216 \\
& Representation Shift   & 80.8 & 26.7 & 3705
                & 26.7 & 28.5 & 3480 \\
& V-Pruner      & 80.8 & 28.3 & 3917
                & 26.6 & 32.5 & 3441 \\
& \textit{Average of baselines}
                & 80.8 & 24.7 & 3314
                & 26.7 & 27.3 & 3010 \\
& \textbf{DORA}
                & \textbf{80.8} & \textbf{21.7} & \textbf{3944}
                & \textbf{26.7} & \textbf{21.0} & \textbf{4233} \\
\midrule

\multirow{10}{*}{ViT-B}
& Native        & 84.6 & 35.2 & 3211
                & 44.0 & 35.2 & 3209 \\
& ToMe          & 83.7 & 25.2 & 2887
                & 43.1 & 29.6 & 2432 \\
& DiffRate      & 83.8 & \textbf{23.1} & 3018
                & --   & --   & -- \\
& PiToMe        & 83.8 & 26.2 & 3494
                & 43.2 & 26.5 & 3458 \\
& MCTF          & 83.8 & 23.7 & 1900
                & 43.2 & 26.6 & 1647 \\
& Zero-TPrune   & 83.8 & 26.8 & 2800
                & 43.2 & 23.3 & 3201 \\
& Representation Shift   & 83.8 & 28.1 & 3508
                & 43.2 & 28.6 & 3457 \\
& V-Pruner      & 83.8 & 29.9 & 3719
                & 43.2 & 34.3 & 3278 \\
& \textit{Average of baselines}
                & 83.8 & 26.1 & 3046
                & 43.2 & 28.2 & 2912 \\
& \textbf{DORA}
                & \textbf{83.8} & 23.6 & \textbf{3725}
                & \textbf{43.2} & \textbf{23.2} & \textbf{3831} \\
\midrule

\multirow{10}{*}{ViT-L}
& Native        & 85.8 & 123.4 & 998
                & 55.3 & 123.4 & 997 \\
& ToMe          & 85.0 & 84.8  & 1013
                & 54.6 & 115.7 & 726 \\
& DiffRate      & 85.1 & 81.8  & 963
                & 54.6 & 77.2  & 1070 \\
& PiToMe        & 85.1 & 89.2  & 1184
                & 54.7 & 102.9 & 1035 \\
& MCTF          & 85.1 & \textbf{73.4}  & 755
                & 54.7 & 109.4 & 464 \\
& Zero-TPrune   & 85.1 & 88.2  & 1159
                & 54.7 & 75.1  & 1304 \\
& Representation Shift   & 85.1 & 86.8  & 1288
                & 54.7 & 94.3  & 1182 \\
& V-Pruner      & 85.1 & 106.7 & 1140
                & 54.6 & 121.5 & 1012 \\
& \textit{Average of baselines}
                & 85.1 & 87.3 & 1071
                & 54.7 & 99.4 & 970 \\
& \textbf{DORA}
                & \textbf{85.1} & 74.9 & \textbf{1394}
                & \textbf{54.7} & \textbf{74.3} & \textbf{1456} \\
\midrule

\multirow{10}{*}{DINOv2}
& Native        & 84.5 & 46.5 & 1842
                & 57.3 & 46.5 & 1842 \\
& ToMe          & 83.6 & 36.2 & 1340
                & 55.9 & 40.8 & 1176 \\
& DiffRate      & --   & --   & --
                & 56.0 & 33.5 & 1640 \\
& PiToMe        & 83.7 & 37.2 & 1854
                & 56.0 & 37.4 & 1843 \\
& MCTF          & 83.7 & 33.8 & 701
                & 56.0 & 36.8 & 628 \\
& Zero-TPrune   & 83.7 & 36.5 & 1085
                & 56.0 & 33.9 & 1282 \\
& Representation Shift   & 83.7 & 37.4 & 2070
                & 56.0 & 41.3 & 1879 \\
& V-Pruner      & 83.7 & 37.5 & 2301
                & 55.9 & 44.0 & 1992 \\
& \textit{Average of baselines}
                & 83.7 & 36.4 & 1559
                & 56.0 & 38.2 & 1491 \\
& \textbf{DORA}
                & \textbf{83.7} & \textbf{30.3} & \textbf{2466}
                & \textbf{56.0} & \textbf{32.5} & \textbf{2389} \\
\bottomrule
\end{tabular*}
\end{table}

\smallhead{ImageNet-1K results.}
At matched accuracy, DORA has the highest measured throughput on every
backbone and, with two small exceptions, also the lowest FLOPs. The
pairwise comparisons show that lower FLOPs do not necessarily imply
higher throughput: on DeiT-B, MCTF nearly matches DORA's computation
(21.9 vs.\ 21.7 GFLOPs) yet reaches only 2{,}122 images/s against
DORA's 3{,}944, while V-Pruner comes within 0.7\% of DORA's throughput
but requires 28.3 GFLOPs to DORA's 21.7. The two exceptions undercut
DORA's computation only marginally---DiffRate on ViT-B by 2.2\% and
MCTF on ViT-L by 2.0\%---while DORA's throughput is 23.4\% and 84.7\%
higher, respectively, showing that FLOPs and wall-clock performance are
not interchangeable. Averaged equally across the four backbones, DORA
reduces FLOPs by 13.2\% and raises throughput by 32.4\% relative to the
baseline means, indicating that the actor and dynamic-sequence
operations do not erase the savings from hard deletion.

\smallhead{Zero-shot transfer to ImageNet-A.}
ImageNet-A consists of naturally occurring adversarial examples with
atypical object appearances, backgrounds, textures, poses, scales, and
partial occlusions, providing a more challenging distribution-shift
setting than ImageNet-1K for preserving recognition accuracy
under token reduction~\citep{hendrycks2021natural}. We transfer the
selected DORA policies directly from ImageNet-1K to ImageNet-A without
retraining, checkpoint re-selection, or any adjustment of the DORA
operating point (Section~\ref{sec:exp-setup}). For accuracy-matched
comparison, each baseline is evaluated at its available ImageNet-A
operating points, and we report the point whose Top-1 accuracy is
closest to that of the transferred DORA policy. Under this protocol,
DORA's advantage widens under distribution shift: averaged across the
four backbones, it uses 20.3\% fewer FLOPs and achieves 45.6\% higher
throughput than the baseline means. DORA also has the lowest FLOPs and
the highest measured throughput among the accuracy-matched operating
points on all four backbones.

\subsection{DORA learns backbone- and input-adaptive pruning}
\label{sec:policy_analysis}

\begin{figure}[t]
    \centering
    \includegraphics[width=\linewidth]{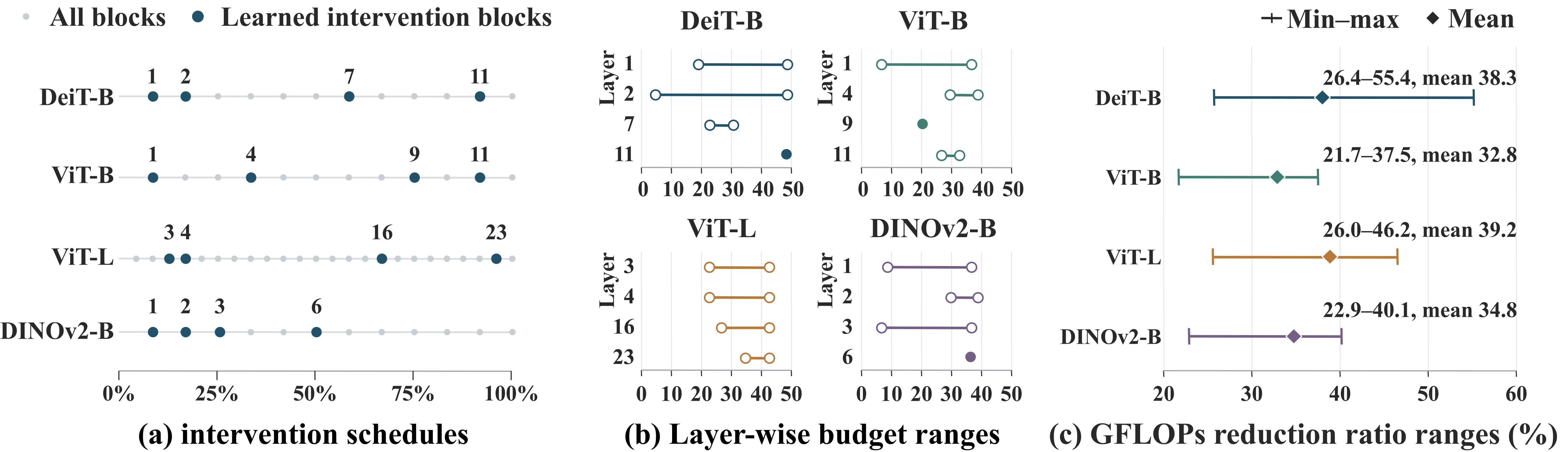}
    \caption{\textbf{Learned adaptivity in where and how much to prune.}
    (a) Intervention locations learned for each backbone. (b) Min--max
    deletion budgets over gate-active validation images at each intervention block;
    filled points denote collapsed single-budget decisions. (c) The
    resulting per-image GFLOPs-reduction ranges and means.}
    \label{fig:gate_budget}
\end{figure}

\smallhead{Where to intervene.}
Figure~\ref{fig:gate_budget} analyzes where DORA intervenes and how much it prunes. 
On the first axis, the gate converges to backbone-specific
intervention schedules rather than inheriting a common hand-designed
placement: DeiT-B intervenes at blocks $\{1,2,7,11\}$, ViT-B at
$\{1,4,9,11\}$, ViT-L at $\{3,4,16,23\}$, and DINOv2-B at
$\{1,2,3,6\}$, where block~3 is conditional and
activates on only 66.5\% of validation images. This learned placement
contrasts with hand-crafted patterns: ToMe merges tokens at every
block~\citep{bolya2023tome}, and Zero-TPrune prunes only at predefined
layers~\citep{wang2024zerotprune},
whereas DORA discovers where to intervene through training, including
input-conditioned activation at individual positions.

\smallhead{How much to prune.}
Several intervention blocks span broad budget
ranges across images, while others converge to narrow or deterministic
budgets (Figure~\ref{fig:gate_budget}(b)). Consequently, per-image
GFLOPs reductions range from 26.4--55.4\% on DeiT-B (mean 38.3\%),
21.7--37.5\% on ViT-B (mean 32.8\%), 26.0--46.2\% on ViT-L (mean
39.2\%), and 22.9--40.1\% on DINOv2-B (mean 34.8\%), in line with the
average reductions in Table~\ref{tab:main_efficiency_joint}. The
qualitative trajectories in Figure~\ref{fig:main_dora} illustrate the
corresponding differences in retained spatial evidence.

\smallhead{Which tokens to prune: selector quality.}
We ask whether the learned selector identifies the
right tokens. Holding DORA's per-image gate decisions and deletion
budgets fixed, we replace only the token-identity rule with random,
similarity-based (ToMe-style), or attention-importance
(Zero-TPrune-style) hard-pruning rules, each operating on its own
evolving features. Under the fixed trajectory, the learned selector
preserves higher Top-1 accuracy: improvements of 1.6--1.8 points over
Random, 0.3--0.9 points over Similarity, and 0.9--3.5 points over
Attention-WPR across the four backbones. This conclusion is conditional
on the jointly learned gate and budget trajectories and does not claim
universal optimality under arbitrary pruning budgets. The controlled
protocol and full results are in Appendix~\ref{app:additional-selector}.

\subsection{Accuracy feedback calibrates fidelity across backbones}
\label{sec:feedback_analysis}

\begin{figure}[t]
    \centering

    \begin{minipage}[t]{0.48\linewidth}
        \vspace{0pt}
        \centering
        \includegraphics[width=\linewidth]{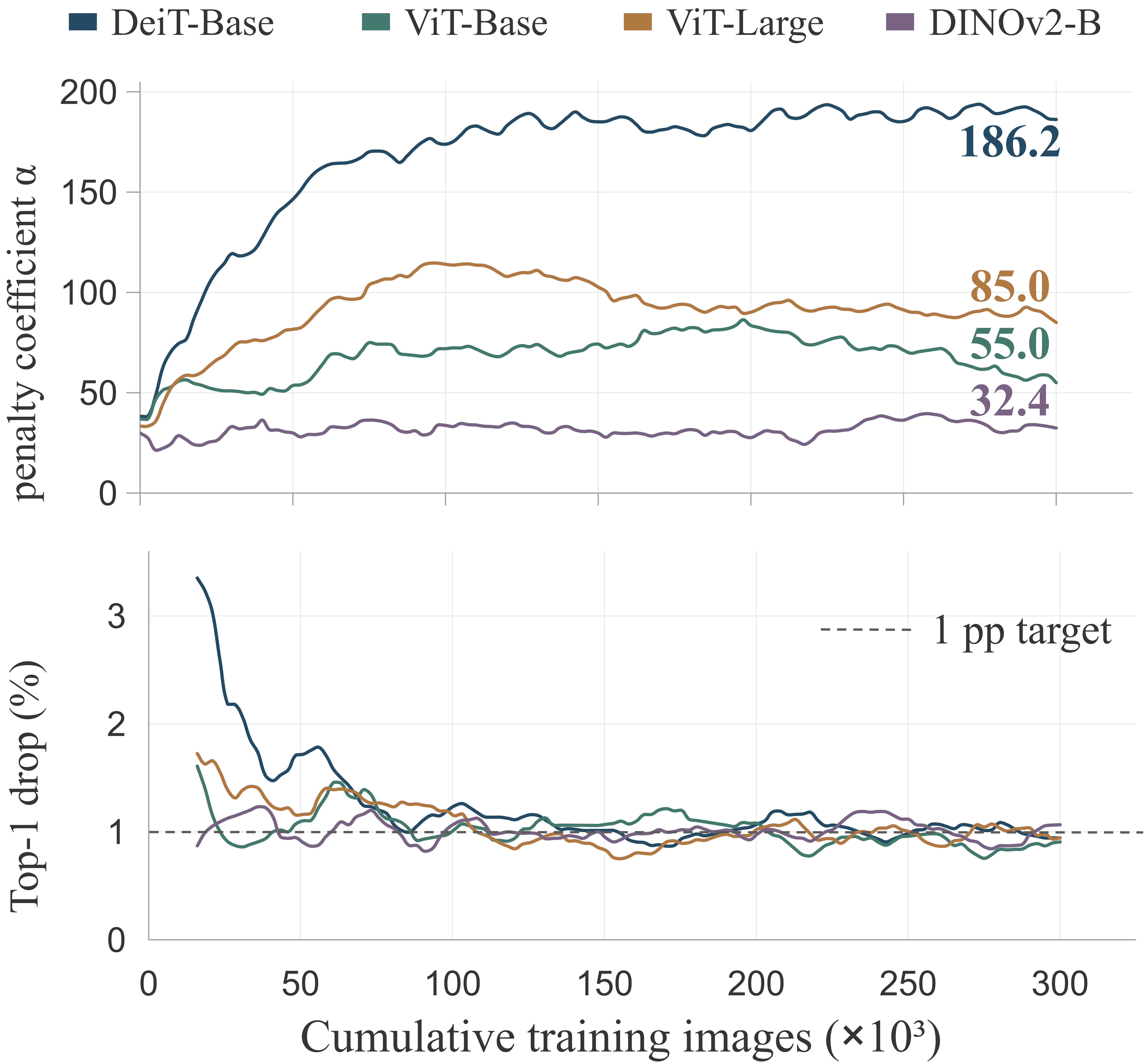}
        \caption{
        \textbf{Closed-loop accuracy feedback.}
        Top: learned fidelity coefficient $a$ across four backbones.
        Bottom: recent eight-check mean Top-1 drop.
        }
        \label{fig:feedback_dynamics}
    \end{minipage}
    \hfill
    \begin{minipage}[t]{0.48\linewidth}
        \vspace{0pt}
        \centering
        \includegraphics[width=\linewidth]{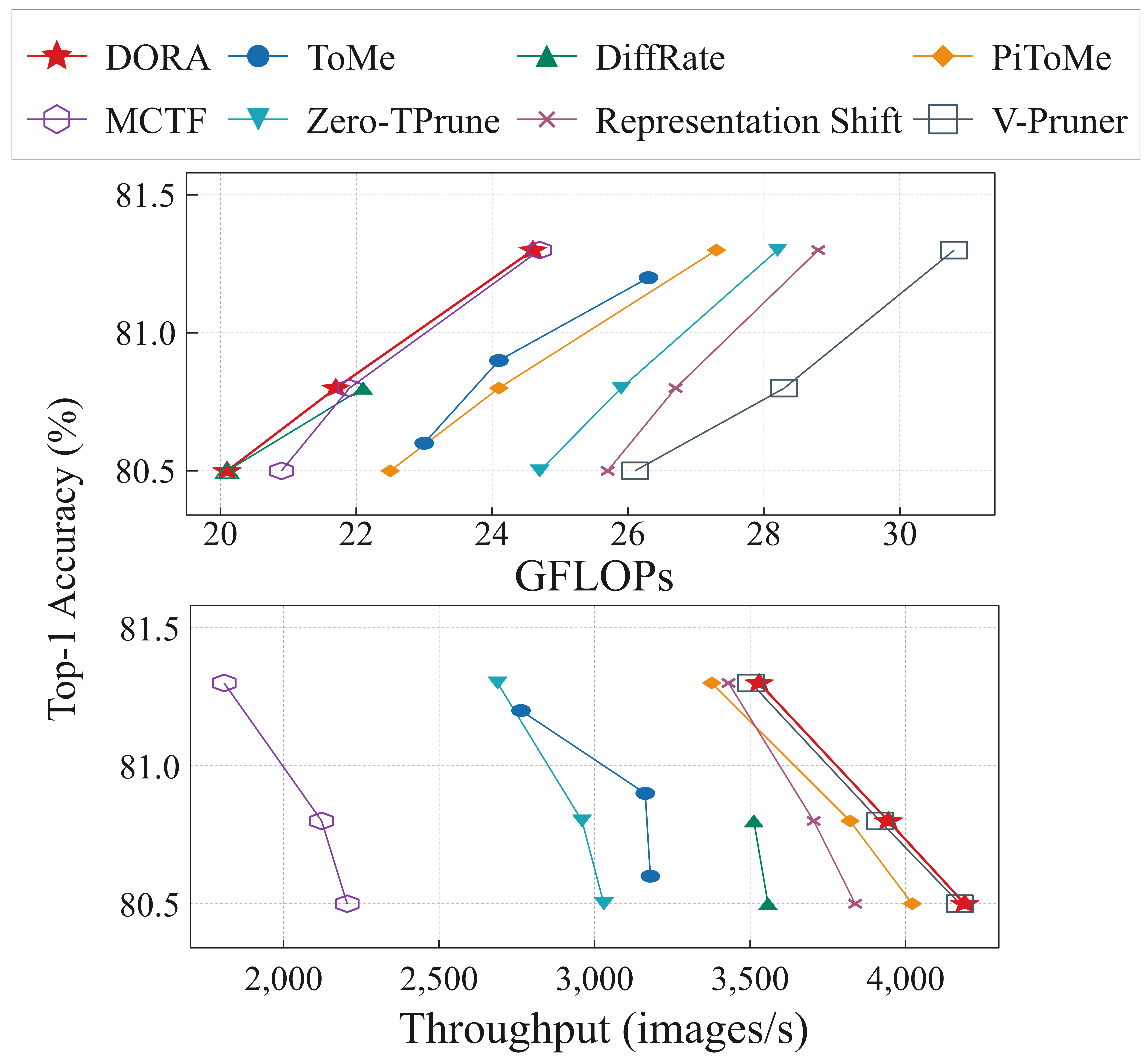}
        \caption{
        \textbf{Feedback-target sweep on DeiT-B.}
        Top: ImageNet-1K Top-1 accuracy versus GFLOPs per image.
        Bottom: Top-1 accuracy versus throughput.
        }
        \label{fig:feedback_tradeoff}
    \end{minipage}

\end{figure}

The same update rule yields different coefficient regimes: by 300K
training images, $a$ reaches 186.2, 55.0, 85.0, and 32.4 for DeiT-B,
ViT-B, ViT-L, and DINOv2-B, respectively
(Figure~\ref{fig:feedback_dynamics}, top). All runs share the
initialization $a=30$, so these endpoints are not backbone-specific
settings; they emerge from the measured behavior of each policy. The
endpoint values describe the ends of the training trajectories and need
not coincide with the coefficients of the checkpoints selected for
final evaluation. At the same time, the recent feedback averages move
toward and then fluctuate around the shared 1-pp target
(Figure~\ref{fig:feedback_dynamics}, bottom). The feedback signal is
measured on a quality-control split that is disjoint from both the PPO
rollout pool and the development split used for checkpoint selection,
so the target is a training signal rather than a guarantee on the final
validation set.

The shared initialization is benign only because the loop is closed.
Fixing $a$ at 30 for the entire run is catastrophic on DeiT-B: no saved
checkpoint satisfies the development-set quality constraint, and the
terminal checkpoint reaches only 78.65\% Top-1 with a 3.14-pp Native
drop (Table~\ref{tab:ablation_stability}). This failure is consistent
with the trajectory in Figure~\ref{fig:feedback_dynamics}: holding the
1-pp target on DeiT-B requires the coefficient to rise sixfold, from 30
to 186.2. A fixed low coefficient under-penalizes fidelity loss and the
policy over-prunes; the closed loop removes this failure mode without
backbone-specific tuning.

The feedback target can also control the operating point. Retraining
DORA on DeiT-B with target drops of 0.5, 1.0, and 1.5 percentage
points produces three distinct accuracy--efficiency regimes without
otherwise changing the method
(Figure~\ref{fig:feedback_tradeoff}): at every target, DORA maintains
competitive GFLOPs and the highest measured throughput among
accuracy-matched methods, occupying the top-left of the
accuracy--GFLOPs plane and the top-right of the accuracy--throughput
plane. The wall-clock advantage thus persists across operating regimes
rather than being tied to the default 1-pp target. Details are given in
Appendix~\ref{app:additional-targets}.

\begin{table}[t]
    \centering
    \caption{
    \textbf{Training-mechanism ablations on DeiT-B.}
    }
    \label{tab:ablation_stability}

    \normalsize
    \renewcommand{\arraystretch}{0.85}
\begin{tabular*}{\linewidth}{
    @{\extracolsep{\fill}}
    lccc
    @{}
}
    \toprule
    Setting &
    \shortstack{Top-1(\%)} &
    GFLOPs &
    \shortstack{Native drop (pp)} \\
    \midrule

    \textbf{Full DORA}
    & \textbf{80.79}
    & \textbf{21.66}
    & \textbf{1.00} \\

    Terminal-only reward
    & 80.81
    & 22.95
    & 0.98 \\

    Fixed coefficient ($a{=}30$)
    & 78.65
    & 17.78
    & 3.14 \\

    \bottomrule
\end{tabular*}
\end{table}

\subsection{Ablations and stability}
\label{sec:ablation_stability}

Replacing localized complete-prefix credit with a terminal-only reward
produces a nearby but weaker selected operating point (80.81\% Top-1,
22.95 GFLOPs versus 80.79\%, 21.66 GFLOPs;
Table~\ref{tab:ablation_stability}), supporting the use of localized
per-step credit. Repeating the full training procedure with three
additional seeds yields closely clustered selected operating points:
$80.82{\pm}0.07$\% Top-1 and $21.76{\pm}0.18$ GFLOPs across four runs,
indicating that the selected accuracy--efficiency regime is stable
across the evaluated training seeds. Full results are reported in
Appendix~\ref{app:additional-seeds}.

\section{Conclusion}
We presented DORA, a reinforcement learning framework that formulates token pruning in vision transformers as a finite-horizon Markov decision process. A hierarchical actor separates whether, how much, and which tokens to prune, while complete-prefix shadow evaluations provide localized credit for these coupled decisions. A privileged critic and a closed-loop accuracy-feedback mechanism stabilize training, and the backbone remains frozen with a deterministic actor at deployment. Across four ViT backbones, DORA consistently improves the accuracy-efficiency trade-off among the evaluated frozen-backbone methods, and hard deletion combined with packed variable-length FlashAttention converts the reduced computation into measured acceleration.

DORA targets throughput-oriented ViT inference where input redundancy varies across images. Our evaluation is limited to image classification on four ViT-type backbones, with runtime benefits measured on GPU execution with variable-length FlashAttention; training also incurs additional cost from shadow evaluations and the privileged critic.

\label{maintextend}

\clearpage
\subsection*{AI use statement}

We used AI assistants only for manuscript drafting and polishing,
first-draft figure generation, discussion of experimental results, and
reference checking. The authors reviewed and edited all AI-assisted
material and take full responsibility for the final paper.

\bibliography{references}
\bibliographystyle{iclr2027_conference}

\clearpage
\appendix

\section{Actor Architecture and Action-Space Details}
\label{app:actor}

\subsection{Gate, shared encoder, budget head, and selector}
\label{app:actor-architecture}

Each backbone has an independently trained actor whose parameters are
shared across blocks, with block-specific learned embeddings. At block
$l\in\{0,\ldots,L-1\}$, the actor reads the backbone keys
$K_l\in\mathbb{R}^{(N_l+1)\times d}$, with attention-head channels
concatenated. Token counts exclude CLS. The history vector is
\begin{equation}
    h_l=\left[\frac{N_l}{N_0},\;\rho_l,\;
    \frac{E_{<l}}{L-1}\right],
    \label{eq:app-actor-history}
\end{equation}
where $\rho_0=0$, $\rho_l=k_{l-1}/N_{l-1}$ for $l>0$, and $E_{<l}$
counts earlier pruning interventions. We take $k_j=0$ at gate-off steps.

The gate concatenates $K_{l,\mathrm{CLS}}$, a learned block embedding
$e_l^G\in\mathbb{R}^{16}$, and $h_l$. A
$(d+19)\!\rightarrow\!64\!\rightarrow\!1$ MLP with a SiLU hidden
activation and sigmoid output produces the activation probability
$p_l^G$. The controller runs only when the gate is active.

The shared encoder constructs
\begin{equation}
    X_{l,j}=W_{\mathrm{in}}K_{l,j}+b_{\mathrm{in}}+e_l^C
    +\mathbf{1}[j=\mathrm{CLS}]W_hh_l,
    \label{eq:app-actor-input}
\end{equation}
where $W_{\mathrm{in}}:d\!\rightarrow\!128$,
$e_l^C\in\mathbb{R}^{128}$, and $W_h:3\!\rightarrow\!128$.
The gate and controller use separate block embeddings. The controller
adds its block embedding to every token and injects history only into
CLS. A single pre-LayerNorm Transformer block, with width 128, four
attention heads, and a $128\!\rightarrow\!256\!\rightarrow\!128$
GELU feed-forward sublayer, produces $Z_l$.

The budget head applies LayerNorm to $Z_{l,\mathrm{CLS}}$ followed by
a $128\!\rightarrow\!128\!\rightarrow\!M$ MLP with GELU,
where $M$ is the output dimension in Table~\ref{tab:app-actor-config}.
Given $k_l$, the selector applies a shared
$128\!\rightarrow\!64\!\rightarrow\!1$ SiLU MLP to
$Z_{l,j}+W_k(k_l/N_l)$, where $W_k:1\!\rightarrow\!128$.
The resulting score $s_{l,j}$ indicates deletion preference; higher
scores favor removal, and CLS is excluded.

\subsection{Backbone-specific budget grids}
\label{app:actor-budgets}

All $L$ blocks are eligible for pruning. The budget grid is
$\mathcal B=\{2,4,\ldots,N_0-4\}$, with $M=(N_0-4)/2$ outputs.
At block $l$, the feasible budgets preserve at least four visual tokens:
\begin{equation}
    \mathcal B_l=\{k\in\mathcal B:k\leq N_l-4\}.
    \label{eq:app-actor-legal-budgets}
\end{equation}
The same rule applies at every depth. No pruning is represented by
gate-off rather than a zero budget; if $\mathcal B_l$ is empty, the
gate remains off.

\begin{table}[!hb]
\centering
\normalsize
\caption{Backbone dimensions and initial budget grids. $M$ is the
budget-head output dimension. Token counts exclude CLS.}
\label{tab:app-actor-config}
\setlength{\tabcolsep}{5pt}
\begin{tabular}{@{}lrrrrl@{}}
\toprule
Backbone & $d$ & $L$ & $N_0$ & $M$ & $\mathcal B_0$ \\
\midrule
DeiT-B/16   & 768  & 12 & 196 & 96  & $\{2,4,\ldots,192\}$ \\
ViT-B/16    & 768  & 12 & 196 & 96  & $\{2,4,\ldots,192\}$ \\
ViT-L/16    & 1024 & 24 & 196 & 96  & $\{2,4,\ldots,192\}$ \\
DINOv2-B/14 & 768  & 12 & 256 & 126 & $\{2,4,\ldots,252\}$ \\
\bottomrule
\end{tabular}
\end{table}

\subsection{Training-time sampling and deterministic inference}
\label{app:actor-sampling}

When $\mathcal B_l\neq\varnothing$, training samples
$g_l\sim\operatorname{Bernoulli}(p_l^G)$. Conditional on $g_l=1$,
$k_l$ is sampled from the categorical distribution
$\pi_B(\cdot\mid o_l)$ over $\mathcal B_l$.
The selector computes its scores once and samples an ordered sequence
$\boldsymbol{i}_l=(i_1,\ldots,i_{k_l})$ without replacement. With
$R_t=\{1,\ldots,N_l\}\setminus\{i_1,\ldots,i_{t-1}\}$, the
Plackett--Luce log-probability is
\begin{equation}
    \log\pi_S(\boldsymbol{i}_l\mid o_l,k_l)
    =\sum_{t=1}^{k_l}\left[
        s_{l,i_t}-\log\sum_{j\in R_t}\exp(s_{l,j})
    \right].
    \label{eq:app-actor-pl}
\end{equation}
Although deletion depends only on the selected set, PPO retains the
sampled order to evaluate the same action's likelihood under the old
and new policies.

At inference, the gate opens when $p_l^G\geq0.5$ and
$\mathcal B_l\neq\varnothing$. The budget selects the highest-logit feasible budget, and the selector removes
the $k_l$ highest-scoring visual tokens. Deletion occurs after the
attention residual and before the MLP, preserving CLS and the relative
order of surviving tokens.
\section{Complete-Prefix Shadow Evaluation and Reward Derivation}
\label{app:reward}

\subsection{Token-layer compression}
\label{app:reward-compression}

Let $L$ be the backbone depth and $N_0$ the initial number of visual
tokens. Token counts exclude CLS, and $k_l=0$ at gate-off steps. 
The token counts before and after deletion satisfy
\begin{equation}
    N_l=N_0-\sum_{j=0}^{l-1}k_j,
    \qquad N_l^{\mathrm{post}}=N_l-k_l.
    \label{eq:app-reward-counts}
\end{equation}
Since deleted tokens remain absent from subsequent blocks, exchanging
the order of summation gives
\begin{equation}
\begin{aligned}
    C
    &=\frac{1}{LN_0}\sum_{l=0}^{L-1}
        \bigl(N_0-N_l^{\mathrm{post}}\bigr) \\
    &=\frac{1}{LN_0}\sum_{l=0}^{L-1}\sum_{j=0}^{l}k_j
      =\sum_{l=0}^{L-1}
        \underbrace{\frac{k_l(L-l)}{LN_0}}_{\Delta C_l}.
\end{aligned}
\label{eq:app-reward-compression}
\end{equation}
The factor $L-l$ counts the post-pruning block outputs in which the
deleted tokens are absent. Because deletion follows the attention
residual and precedes the MLP, these tokens still participate in
current-block attention. Thus $C$ serves as a normalized token-layer
training surrogate.

\subsection{Shadow evaluation procedure}
\label{app:reward-shadow}

For each image, let $z^N$ be the logits of the unpruned Native model.
A complete-prefix shadow produces $z^{(l)}$ by replaying the sampled
rollout's recorded deletions at blocks $0,\ldots,l$ and completing
all remaining blocks without further pruning. No actor decisions
are resampled. Each shadow uses the same frozen model and preprocessing as the Native reference.

Define $p^N=\operatorname{softmax}(z^N/\tau)$ and
$p^{(l)}=\operatorname{softmax}(z^{(l)}/\tau)$. The per-image fidelity
cost is
\begin{equation}
    D_l=\tau^2\sum_c p_c^N
        \log\frac{p_c^N}{p_c^{(l)}},
    \qquad \tau=3.
    \label{eq:app-reward-kd}
\end{equation}
The empty prefix is the Native forward, giving $D_{-1}=0$.
A gate-off step introduces no additional pruning, so
$D_l=D_{l-1}$ when $k_l=0$; the full prefix reproduces the rollout's
final prediction. Shadow evaluation and reward construction are
gradient-free.

\subsection{Telescoping reward derivation}
\label{app:reward-telescoping}

The coefficient $a_u$ remains fixed throughout a rollout and its
associated PPO update, with feedback changes applied to subsequent
rollouts. The local reward is
\begin{equation}
    r_l=25\Delta C_l-a_u(D_l-D_{l-1}).
    \label{eq:app-reward-step}
\end{equation}
For the undiscounted objective ($\gamma=1$),
Eq.~\eqref{eq:app-reward-compression} and cancellation of adjacent
fidelity terms yield
\begin{equation}
\begin{aligned}
    \sum_{l=0}^{L-1}r_l
    &=25C-a_u\sum_{l=0}^{L-1}(D_l-D_{l-1}) \\
    &=25C-a_u\bigl(D_{L-1}-D_{-1}\bigr) \\
    &=25C-a_uD_{\mathrm{final}},
\end{aligned}
\label{eq:app-reward-telescope}
\end{equation}
where $D_{\mathrm{final}}=D_{L-1}$ is the complete rollout's fidelity
cost. The signed increment $D_l-D_{l-1}$ measures the current action's
marginal effect on final-prediction fidelity, rather than repeatedly
penalizing previously accumulated divergence. No additional terminal
reward is added, since the final fidelity cost is already included.
\section{Closed-Loop Accuracy Feedback}
\label{app:feedback}

\subsection{Feedback split and cadence}
\label{app:feedback-cadence}

We reserve 16,384 ImageNet-1K training images as a feedback split,
disjoint from the PPO rollout pool and development split
(Appendix~\ref{app:experiments-data}). The feedback split 
is used only to update the fidelity coefficient;
development and validation data are not used for feedback.

The QC split contains eight fixed shards of 2,048 images, evaluated
in round-robin order. Each rollout contains 256 images, 
and feedback is evaluated every eight PPO updates.
Each check uses one shard and the deterministic actor defined in
Appendix~\ref{app:actor-sampling}, after rollout collection and before
the corresponding policy update.

\subsection{Coefficient initialization and update}
\label{app:feedback-update}

Let $u=1,2,\ldots$ index PPO updates, and let $a_u$ be the fidelity
coefficient used in update $u$. All four backbones use $a_1=30$.
At feedback checks, $d_u$ denotes the current shard's paired Top-1
drop as an accuracy fraction. For target $d^\star$, the update is
\begin{equation}
    a_{u+1}=
    \begin{cases}
    \operatorname{clip}_{[0,300]}\!\left(
        a_u+5\operatorname{clip}_{[-2,2]}\!\left(
            \dfrac{d_u-d^\star}{0.01}
        \right)\right),
        & u\equiv0\pmod{8},\\
    a_u, & \text{otherwise}.
    \end{cases}
    \label{eq:app-feedback-update}
\end{equation}
The default target is $d^\star=0.01$, corresponding to one percentage
point. The target sweep in Appendix~\ref{app:additional-targets} uses
$d^\star\in\{0.005,0.01,0.015\}$; only the target changes, while the
normalization scale remains 0.01 and all other feedback parameters
remain fixed.

The coefficient $a_u$ remains fixed throughout rollout collection and
the associated PPO update (Appendix~\ref{app:reward}); $a_{u+1}$ is
applied from the next rollout onward. The target regulates training
but does not guarantee the same accuracy drop on development or
validation data.

\subsection{Paired accuracy-drop computation}
\label{app:feedback-paired-drop}

Native and DORA predictions are compared on the same QC images,
using the same frozen backbone weights and input preprocessing.
For a shard of $M=2{,}048$ images with labels $y_i$, let
$c_i^N=\mathbf{1}[\hat y_i^N=y_i]$ and
$c_i^D=\mathbf{1}[\hat y_i^D=y_i]$ indicate correct Native and DORA
Top-1 predictions. The paired drop is
\begin{equation}
    d_u=\frac{1}{M}\sum_{i=1}^{M}(c_i^N-c_i^D)
       =\operatorname{Acc}_N-\operatorname{Acc}_D,
    \label{eq:app-feedback-paired-drop}
\end{equation}
where both accuracies are fractions evaluated on that shard.
Equivalently, $d_u$ is the fraction of Native-correct/DORA-wrong
predictions minus the fraction of Native-wrong/DORA-correct
predictions. It retains its sign, and $100d_u$ gives the drop in
percentage points.
\section{Privileged Critic and PPO Optimization}
\label{app:critic}

\subsection{Critic inputs and architecture}
\label{app:critic-architecture}

The critic is used only during training. At block
$l\in\{0,\ldots,L-1\}$, it reads the pre-action residual tokens
$H_l$, the Native logits $z^N\in\mathbb R^{1000}$ of the same image,
and a compact scalar state. Let
$p^N=\operatorname{softmax}(z^N)$,
$m^N=p^N_{(1)}-p^N_{(2)}$, and
\[
    \phi^N=
    \left[
        p^N_{(1)},\;
        m^N,\;
        \mathcal H(p^N),\;
        \mathbf1[m^N\leq\tau_b]
    \right],
\]
where $p^N_{(1)}$ and $p^N_{(2)}$ are the two largest Native
probabilities and $\tau_b$ is a fixed backbone-specific low-margin
threshold. The scalar input is
\begin{equation}
    q_l=
    \left[
        \frac{N_l}{N_0},\;
        \frac{l}{L-1},\;
        C_{<l},\;
        \frac{E_{<l}}{L-1},\;
        D_{l-1},\;
        \phi^N,\;
        \frac{a_u}{300}
    \right],
    \label{eq:app-critic-scalars}
\end{equation}
where $C_{<l}=\sum_{j<l}\Delta C_j$, $E_{<l}$ counts earlier
pruning interventions, and $D_{-1}=0$ as defined in
Appendix~\ref{app:reward}.

The token branch projects $H_l$ to width 256, adds learned block and
CLS/visual-type embeddings, prepends a learned value token, and
processes the sequence with two pre-LayerNorm Transformer blocks
(width 256, four heads, feed-forward width 512). The value-token
representation is concatenated with separate 64-dimensional
projections of $q_l$ and $z^N$, yielding
$u_l\in\mathbb R^{384}$.

Independent $384\!\rightarrow\!128\!\rightarrow\!1$ MLPs predict
the gate and budget values $V_l^G$ and $V_l^B$. The selector value
additionally conditions on the chosen budget:
\[
    V_l^S
    =
    f_S\!\left(
        [u_l;\,e(k_l);\,k_l/N_l]
    \right),
\]
where $e(k_l)\in\mathbb R^{32}$ is a learned budget embedding and
$f_S$ is a $417\!\rightarrow\!128\!\rightarrow\!1$ MLP.
No value head observes selected token identities, post-action states,
or ground-truth labels.

\subsection{Value targets and PPO optimization}
\label{app:critic-targets}

For each $L$-step trajectory, we use GAE with
$\gamma=1$ and $\lambda_{\mathrm{GAE}}=0.95$.
A common $\lambda$-return is constructed from the pre-update gate
values:
\begin{equation}
\begin{aligned}
    \delta_l
    &=r_l+V_{l+1}^{G,\mathrm{old}}-V_l^{G,\mathrm{old}},\\
    A_l^G
    &=\delta_l+\lambda_{\mathrm{GAE}}A_{l+1}^G,\\
    \widehat R_l^\lambda
    &=A_l^G+V_l^{G,\mathrm{old}},\\
    A_l^X
    &=\widehat R_l^\lambda-V_l^{X,\mathrm{old}},
    \qquad X\in\{G,B,S\},
\end{aligned}
\label{eq:app-critic-gae}
\end{equation}
with $V_L^{G,\mathrm{old}}=A_L^G=0$.
Thus all three decision types share the same return while using
their own value baselines.

Gate, budget, and selector are optimized with separate clipped PPO
objectives with clip parameter $\epsilon=0.2$. Their advantages are
standardized separately over the corresponding eligible rollout
transitions. Gate decisions require a nonempty feasible budget set,
while budget and selector objectives are defined only for gate-on
transitions. All steps contribute to GAE and $V_G$ regression;
$V_B$ and $V_S$ are trained only on gate-on transitions.

The three value heads regress the shared $\lambda$-return using
running-return normalization and a PPO-style clipped Huber loss.
Normalization is used only for value regression and does not alter
the rewards or GAE computation.

\subsection{PCGrad}
\label{app:critic-pcgrad}

Budget and selector share the controller encoder and can induce
conflicting gradients. Let $g_B$ and $g_S$ denote the gradients of
their respective PPO losses, including entropy regularization, with
respect to the shared controller parameters, and define
$c=\min(g_B^\top g_S,0)$. We use symmetric PCGrad:
\begin{equation}
    \widetilde g_B
    =
    g_B-\frac{c\,g_S}{\|g_S\|_2^2+10^{-12}},
    \qquad
    \widetilde g_S
    =
    g_S-\frac{c\,g_B}{\|g_B\|_2^2+10^{-12}}.
    \label{eq:app-critic-pcgrad}
\end{equation}
The projected gradients are summed on the shared controller, while
the budget and selector heads retain their original private
gradients. The resulting controller gradient is clipped to norm
0.5 before optimization. Gate and critic updates do not use PCGrad.

\subsection{Training hyperparameters}
\label{app:critic-hyperparameters}

The optimization settings in Table~\ref{tab:app-critic-training}
are shared across all four backbones.

\begin{table}[!htbp]
\centering
\normalsize
\caption{Shared optimization settings for DORA training.}
\label{tab:app-critic-training}
\setlength{\tabcolsep}{4pt}
\begin{tabular}{@{}p{0.38\linewidth}p{0.58\linewidth}@{}}
\toprule
Setting & Value \\
\midrule
Rollout / total training &
256 images / 299,776 images (1,171 updates) \\

Actor / critic epochs &
4 / 4 per update \\

Gate / controller / critic minibatch &
512 / 64 / 128 transitions \\

Optimizer / gradient clipping &
Separate Adam optimizers / max norm 0.5 \\

Gate / critic learning rate &
$5\times10^{-5}$ / $10^{-4}$ \\

Controller learning rates &
Shared: $2\times10^{-5}$;
budget: $5\times10^{-5}$;
selector: $2\times10^{-5}$ \\

Learning-rate schedule / main-run seed &
Constant, no warm-up / 42 \\
\bottomrule
\end{tabular}
\end{table}
\section{Experimental Protocol and Implementation Details}
\label{app:experiments}

\subsection{Data splits}
\label{app:experiments-data}

We use mutually disjoint ImageNet-1K training subsets of 299,776,
16,384, and 10,240 images for PPO rollouts, accuracy feedback
(Appendix~\ref{app:feedback}), and checkpoint selection, respectively.
Final accuracy is evaluated on all 50,000 ImageNet-1K validation
images and all 7,500 ImageNet-A images. ImageNet-A predictions are
restricted to its official 200-class mapping.

All splits use deterministic bicubic resizing and $224\times224$
center cropping, without stochastic augmentation. The resize short
side is 248 for ViT-L/16 and 256 otherwise. ViT-B/16 and ViT-L/16
use RGB mean and standard deviation $(0.5,0.5,0.5)$; DeiT-B/16 and
DINOv2-B/14 use mean $(0.485,0.456,0.406)$ and standard deviation
$(0.229,0.224,0.225)$.

\subsection{Checkpoint selection}
\label{app:experiments-selection}

Let $\mathcal C$ be the checkpoints saved during training. The default
checkpoint minimizes development-set GFLOPs subject to a Native-relative
Top-1 drop of at most one percentage point:
\begin{equation}
    c^\star\in\operatorname*{arg\,min}_{c\in\mathcal C:\,
        A_{\mathrm N}^{\mathrm{dev}}-A_c^{\mathrm{dev}}\leq1.0}
        F_c^{\mathrm{dev}},
    \label{eq:app-experiments-selection}
\end{equation}
where accuracies are percentages and $F_c^{\mathrm{dev}}$ is mean
GFLOPs per image. Neither validation accuracy nor throughput is used
for DORA checkpoint selection. The selected checkpoint transfers to
ImageNet-A unchanged. If an ablation has no feasible checkpoint, its
final checkpoint is reported as constraint-violating.

Baseline operating points are matched separately on each dataset:
we select the measured point closest to DORA's Top-1 within
0.1 percentage points, without interpolation. Missing matches are
omitted. Baseline means are unweighted over available matching
methods, excluding Native and DORA.

\subsection{Baseline implementations}
\label{app:experiments-baselines}

All methods share the frozen backbone weights, classifier, and
preprocessing within each backbone setting. We use non-distilled
DeiT-B/16, AugReg ViT-B/16 and ViT-L/16, and register-free
DINOv2-B/14 with its official LC1 readout. The latter preserves
LayerScale and concatenates normalized CLS with the mean of normalized
surviving visual tokens.

Zero-TPrune is a paper-based reimplementation. DiffRate uses author
schedules as candidates, including ViT-L-MAE schedules transferred
to ViT-L and DeiT-B schedules scaled by $256/196$ for DINOv2 token
counts. All candidate GFLOPs are recomputed under the common accounting
below. V-Pruner's offline mask search is repeated independently for
each backbone and budget; its exported mask is fixed across images
and applied before the first block. Offline search is excluded from
deployment measurements.

\subsection{GFLOPs accounting}
\label{app:experiments-flops}

We count one multiply--accumulate as two FLOPs and include the complete
online forward: patch embedding, backbone, classifier, and each method's
token-reduction computations, including DORA's actor. With $F_i$ denoting
an image's floating-point operation count, the dataset average is
\begin{equation}
    F_{\mathcal D}=\frac{1}{10^9|\mathcal D|}
        \sum_{i\in\mathcal D}F_i
    \quad\text{GFLOPs/image}.
    \label{eq:app-experiments-flops}
\end{equation}
Counts follow each image's realized token trajectory before averaging.
At a DORA intervention block, attention uses $N_l+1$ tokens and the
MLP uses $N_l-k_l+1$, including CLS. Non-floating-point selection and
memory operations contribute to runtime rather than GFLOPs.
Training-only critic and shadow computations are excluded.

\subsection{Runtime and throughput measurement}
\label{app:experiments-runtime}

Timing uses the same RTX~4090, PyTorch 2.11, and CUDA 12.6,
with batch size 320 and BF16 matrix operations. Score comparisons
use FP32; TF32 and \texttt{torch.compile} are disabled. 
In our evaluated implementations, DORA uses packed variable-length
FlashAttention, while Native, PiToMe, Representation Shift, and
V-Pruner also use FlashAttention. Zero-TPrune uses dense attention
in blocks requiring explicit attention graphs and FlashAttention
in the remaining blocks. ToMe, MCTF, and DiffRate retain dense
attention to preserve their proportional-attention and token-size
weighting operations.

ImageNet-1K timing uses 8,000 GPU-resident, preprocessed validation
images in 25 batches. Each of five repeats cycles through these
batches for 200 warm-up and 200 timed forwards, with CUDA
synchronization immediately before and after the timed loop. For
elapsed wall time $T_r$ in seconds, repeat $r$ yields
\begin{equation}
    t_r=\frac{1000T_r}{200}\ \text{ms/batch},
    \qquad q_r=\frac{320\cdot200}{T_r}\ \text{images/s}.
    \label{eq:app-experiments-timing}
\end{equation}
We report the median latency and throughput across repeats.

For ImageNet-A, we use a fixed random
subset of 7,360 images, arranged into 23 full batches of
320 without padding. The same GPU-resident procedure, warm-up,
repeats, and aggregation apply.

Timed forwards include the model, online token selection,
merging/pruning, packing, and compaction. DORA recomputes its decisions
on every forward. Image decoding, preprocessing, host-to-device transfer,
and offline training or search are excluded.
\section{Additional Ablations and Analyses}
\label{app:additional}

\subsection{Selector substitution study}
\label{app:additional-selector}

We fix DORA's per-image gate decisions and deletion budgets and
replace only token selection. \emph{Random} samples $k_l$ visual
tokens uniformly; \emph{Similarity} removes the $k_l$ most redundant
tokens using a ToMe-style key-similarity rule~\citep{bolya2023tome};
and \emph{Attention-WPR} removes the $k_l$ lowest-importance tokens
under a Zero-TPrune-style attention graph~\citep{wang2024zerotprune}.
Each substitute uses its own evolving pruned representation while
following the recorded gate and budget decisions. These are component
substitutions, not evaluations of the complete baseline methods.

\begin{table}[!ht]
\centering
\normalsize
\caption{Selector substitution on ImageNet-1K (Top-1, \%). Random
reports mean $\pm$ standard deviation over five sampling seeds.
Values are rounded to one decimal place.}
\label{tab:selector}
\setlength{\tabcolsep}{4.5pt}
\begin{tabular}{@{}lcccc@{}}
\toprule
Backbone & DORA selector & Random & Similarity & Attention-WPR \\
\midrule
DeiT-B/16   & \textbf{80.8} & $79.1\pm0.1$ & 80.5 & 77.3 \\
ViT-B/16    & \textbf{83.8} & $82.0\pm0.2$ & 83.1 & 81.6 \\
ViT-L/16    & \textbf{85.1} & $83.3\pm0.0$ & 84.2 & 84.2 \\
DINOv2-B/14 & \textbf{83.7} & $82.1\pm0.1$ & 83.3 & 82.2 \\
\bottomrule
\end{tabular}
\end{table}

The learned selector preserves higher Top-1 accuracy than all three
substitutes on every backbone (Table~\ref{tab:selector}). This result
is conditional on DORA's learned gate and budget decisions.

\subsection{Accuracy-feedback target sweep}
\label{app:additional-targets}

We train DORA on DeiT-B with feedback targets
$d^\star\in\{0.005,0.01,0.015\}$, corresponding to accuracy drops
of 0.5, 1.0, and 1.5 percentage points. Only $d^\star$ changes: the
feedback normalization scale remains 0.01 and all other training
settings are fixed
(Appendices~\ref{app:actor}--\ref{app:critic}). Each operating
point comes from a separate training run.

For each run, we select the lowest-GFLOPs checkpoint satisfying the
corresponding Native-relative development-set accuracy constraint,
following Appendix~\ref{app:experiments-selection}. Baselines are
matched to each DORA point's realized validation accuracy within 0.1
percentage points. The resulting operating points are shown in
Figure~\ref{fig:feedback_tradeoff}
(Section~\ref{sec:feedback_analysis}).

\subsection{Four-seed stability}
\label{app:additional-seeds}

We repeat DeiT-B training with three additional seeds, keeping the
data and optimization protocol fixed. Each run's checkpoint is
selected independently under the 1-pp development-set constraint in
Eq.~\eqref{eq:app-experiments-selection}.

\begin{table}[!ht]
\centering
\normalsize
\caption{Four-seed DeiT-B results on ImageNet-1K validation.
Checkpoints are selected on the development split. The final row
reports mean $\pm$ sample standard deviation across the four
training runs.}
\label{tab:seed-stability}
\setlength{\tabcolsep}{8pt}
\begin{tabular}{@{}lccc@{}}
\toprule
Run & Top-1 (\%) & GFLOPs/image & Native drop (pp) \\
\midrule
Original seed     & 80.79 & 21.66 & 1.00 \\
Additional seed 1 & 80.83 & 21.90 & 0.96 \\
Additional seed 2 & 80.75 & 21.57 & 1.04 \\
Additional seed 3 & 80.90 & 21.93 & 0.89 \\
\midrule
Mean $\pm$ std. &
$80.82\pm0.07$ &
$21.76\pm0.18$ &
$0.97\pm0.07$ \\
\bottomrule
\end{tabular}
\end{table}

The selected operating points remain closely clustered across the
four runs (Table~\ref{tab:seed-stability}), with
$80.82\pm0.07\%$ Top-1 and $21.76\pm0.18$ GFLOPs per image.
The development-set constraint is applied independently to each run;
the reported Native drops are measured on ImageNet-1K validation.

\end{document}